\documentclass{article}
\usepackage{ijcai25}

\usepackage{times}
\usepackage{soul}
\usepackage{url}
\usepackage[hidelinks]{hyperref}
\usepackage[utf8]{inputenc}
\usepackage[small]{caption}
\usepackage{graphicx}
\usepackage{amsmath}
\usepackage{amsthm}
\usepackage{booktabs}
\usepackage{algorithm}
\usepackage{algorithmic}
\usepackage[switch]{lineno}

\title{A Smart Multimodal Healthcare Copilot with Powerful LLM Reasoning}

\author{
Xuejiao Zhao$^{1,2\ast}$\and
Siyan Liu$^{1,2}$\thanks{Equal Contributions}\and
Su-Yin Yang$^{3,4}$\and
Chunyan Miao$^{1,2}$\thanks{Corresponding Authors}
\affiliations
$^1$Joint NTU-UBC Research Centre of Excellence in Active Living for the Elderly (LILY), NTU \\
$^2$College of Computing and Data Science, Nanyang Technological University (NTU), Singapore\\ 
$^3$Tan Tock Seng Hospital, Singapore 
$^4$Woodlands Health, Singapore 
\emails
\{xjzhao,siyan.liu,ascymiao\}@ntu.edu.sg,
 su\_yin\_yang@wh.com.sg
}

\begin{document}

\maketitle

\begin{abstract}
    Misdiagnosis causes significant harm to healthcare systems worldwide, leading to increased costs and patient risks. MedRAG is a smart multimodal healthcare copilot equipped with powerful large language model (LLM) reasoning, designed to enhance medical decision-making. It supports multiple input modalities, including non-intrusive voice monitoring, general medical queries, and electronic health records. MedRAG provides diagnostic, treatment, medication, and follow-up questioning recommendations. Leveraging retrieval-augmented generation enhanced by knowledge graph-elicited reasoning, it retrieves and integrates critical diagnostic insights, reducing the risk of misdiagnosis. MedRAG is evaluated on public and private datasets, outperforming existing models and offering more specific and accurate healthcare assistance. The MedRAG demonstration video is available at \textit{\url{https://www.youtube.com/watch?v=PNIBDMYRfDM}}. The code is available at \textit{\url{https://github.com/SNOWTEAM2023/MedRAG}}
\end{abstract}

\section{Introduction}
Misdiagnosis remains a critical challenge in healthcare, leading to significant patient harm and increased healthcare costs~\cite{newman2024burden,dixitelectronic}. In clinical practice, decision-making relies on integrating diverse information sources, yet existing AI-assisted diagnostic systems struggle to effectively process and reason across multiple modalities~\cite{lee2021survey,avanade_health_ai_copilot_2024,rao2024survey}. To address this, we present MedRAG~\cite{zhao2025medrag}, a smart multimodal healthcare copilot equipped with powerful large language model (LLM) reasoning, designed to enhance medical decision-making through multimodal integration and knowledge graph (KG)-elicited reasoning.

Through interviews with healthcare professionals, we identified key requirements for an effective AI-driven diagnostic assistant. Doctors emphasized that an ideal system should incorporate three primary input modalities to comprehensively support clinical workflows~\cite{openai_color_health_2023,amballa2023ai,zakka2024almanac,wei2018comprehensive}:

\begin{itemize}
    \item Non-intrusive voice monitoring – Seamlessly captures real-time doctor-patient conversations during consultations without disruption. This enables instant follow-up questioning and context-aware diagnostic recommendations, enhancing decision-making efficiency without diverting attention.~\cite{wsj_openai_healthcare_2023,ren2024healthcare}.
    \item General medical queries – Allows doctors to interactively refine differential diagnoses, seek clarifications, and receive personalized treatment suggestions in real time. This serves as an intelligent assistant for both clinical and patient-facing decision support.
    \item Electronic Health Records (EHRs) – Analyzes similar cases to provide reasoning-enhanced diagnostics and personalized treatment recommendations, ensuring data-driven support for complex decisions.

\end{itemize}

While retrieval-augmented generation (RAG) has been proposed for medical AI applications, existing heuristic-based RAG models often fail to differentiate between diseases with similar manifestations~\cite{wu2024medical,guu2020retrieval,edge2024local}. Doctors noted that these models tend to generate vague or incorrect recommendations, lacking structured reasoning capabilities~\cite{zelin2024rare,li2023chatdoctor,wu2024pmc}. To overcome this limitation, we introduce KG-elicited reasoning, a key technology in MedRAG that enhances diagnostic accuracy by integrating structured medical knowledge with patient data.

MedRAG systematically constructs a hierarchical diagnostic KG, capturing subtle yet critical diagnostic differences. This KG is dynamically queried based on patient-specific manifestations and integrated with retrieved EHRs, allowing the system to reason through uncertainties and generate precise, context-aware diagnostic suggestions. Additionally, MedRAG proactively proposes follow-up questions to refine ambiguous cases, further supporting clinical workflows.

We evaluate MedRAG on both public (DDXPlus) and private (CPDD) datasets collected from Tan Tock Seng Hospital of Singapore. The results demonstrate its superiority over existing RAG approaches in diagnostic accuracy, specificity, and reasoning-based decision support. Our demo highlights how KG-elicited reasoning transforms MedRAG into a powerful, intelligent, and adaptable healthcare copilot, capable of assisting doctors across diverse clinical scenarios.

\begin{figure*}[htbp] 
    \centering
    \label{fig:framework}
    \includegraphics[width=1\textwidth]{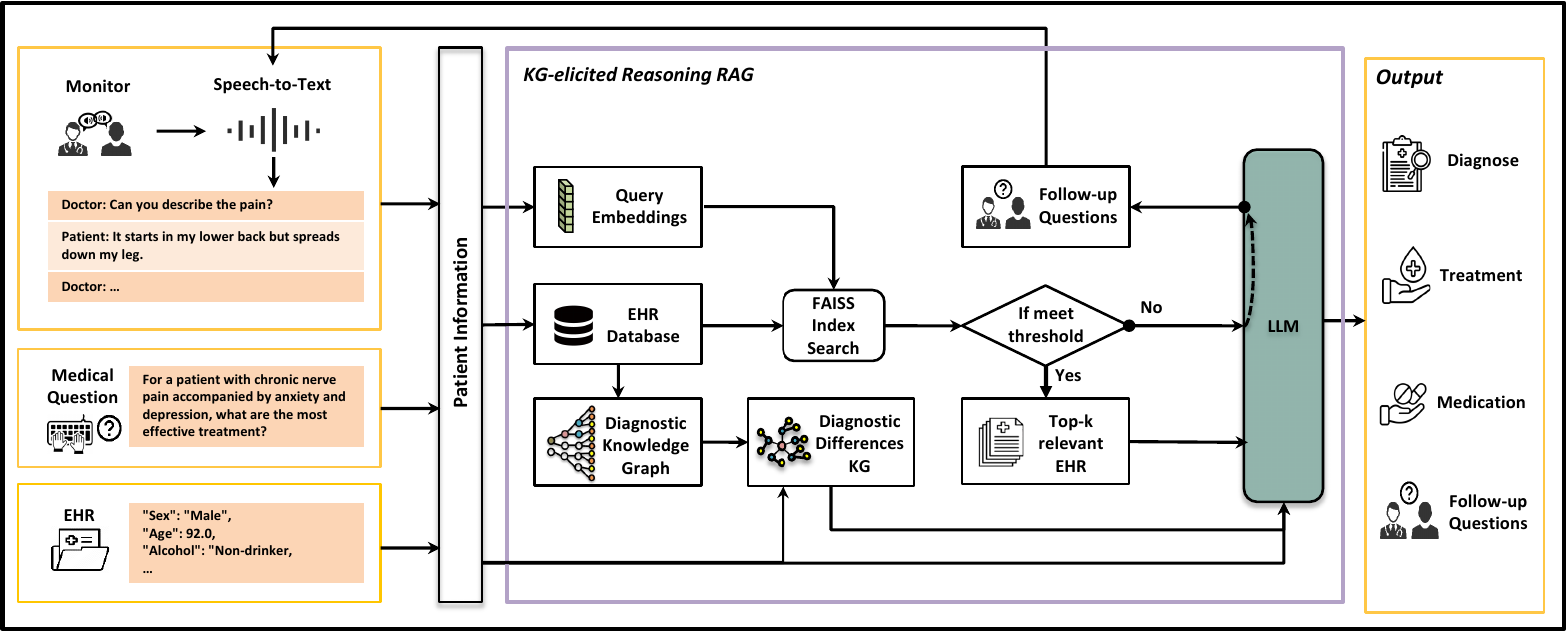} 
    \caption{Framework of Multimodal Healthcare Copilot-MedRAG} 
    \label{fig:framework} 
\end{figure*}
\vspace{-7pt}

\section{System Design}

As shown in Figure \ref{fig:framework}, MedRAG incorporates three different modes of input, a KG-elicited reasoning RAG module and four outputs. MedRAG can seamlessly support various open-source and close-source LLMs, ensuring high adaptability and easy deployment in medical settings.

\subsection{Multimodal Input}
MedRAG provides three input modes to accommodate different clinical scenarios, as illustrated in Figure~\ref{fig:framework}. MedRAG monitors doctor-patient conversations during consultations without interruption using Google's Speech-to-Text API in real time~\cite{GoogleSTT}. With one-click activation at the beginning of the consultation, MedRAG automatically handles information collection, analysis, follow-up questions and diagnostic recommendations, reducing doctors' cognitive and operational workload while allowing them to focus on patient interaction. Doctors can also upload existing files like undiagnosed EHRs or simply type questions by keyboard. All collected information is processed by the KG-elicited Reasoning RAG for further diagnostic analysis.

\subsection{Knowledge Graph-elicited Reasoning RAG}
Knowledge graph-elicited reasoning RAG serves as the core analytical module of MedRAG, it constructs a diagnostic KG from the existing EHR database and identifies the most relevant subgraph based on the patient’s manifestations. It elicits the reasoning ability of the RAG model by extracting relevant triplets as context, which are then fed to the backbone LLM along with retrieved relevant documents, enabling more accurate and structured diagnostic insights.

\subsubsection{Diagnostic Knowledge Graph}
Given the EHR database, MedRAG constructs a four-tiered diagnostic KG by clustering diseases with similar manifestations into hierarchical categories while manifestations of each disease are decomposed into unique features~\cite{zhao2017hdskg}. Features, diseases, subcategories, and categories are structured as nodes to form an undirected KG. Further, we apply GPT-4o to augment the differentiation of similar diseases by expanding more unique features of each disease within each subcategory. Given an undiagnosed patient’s manifestations, MedRAG identifies the most relevant subcategory, and triplets $\langle \text{disease, relation, feature} \rangle$ associated with the identified subcategory are gathered as contextual information to elicit the reasoning capability of the backbone LLM.

\subsubsection{Retrieval-Augmented Generation}
To provide backbone LLM with case-specific information and mitigate hallucinations in generated outputs, we apply the RAG method, retrieving relevant documents before generation. In MedRAG, we use the EHR database as retrieval documents, as EHRs are systematically collected and structured within hospital databases. When patients' disease manifestations are fed into MedRAG, the system measures the semantic similarity between input information and EHRs using cosine similarity. The top 3 relevant EHRs are then selected to provide the contextual input for backbone LLM. OpenAI’s text-embedding-3-large API is used as the text encoder to generate embeddings for both input information and EHRs.

\subsubsection{Proactive Question Generation}
When monitoring a medical consultation, MedRAG evaluates whether sufficient information is available for diagnostic reasoning by analyzing the semantic similarity of the input data and determining whether some EHRs meet a predefined threshold. If it is insufficient, MedRAG identifies the most critical unmentioned disease features in the diagnostic KG to differentiate between similar diseases and formulates follow-up questions. Otherwise, MedRAG proceeds to generate diagnostic recommendations.

\section{User Interface and Evaluation}
We present the user interface (UI) and evaluation, including a case study to showcase MedRAG's performance and a demonstration scenario to illustrate its user interaction and diagnostic support for doctors.

\subsection{UI of MedRAG}
\begin{figure}[h!] 
    \centering
    \includegraphics[width=\linewidth]{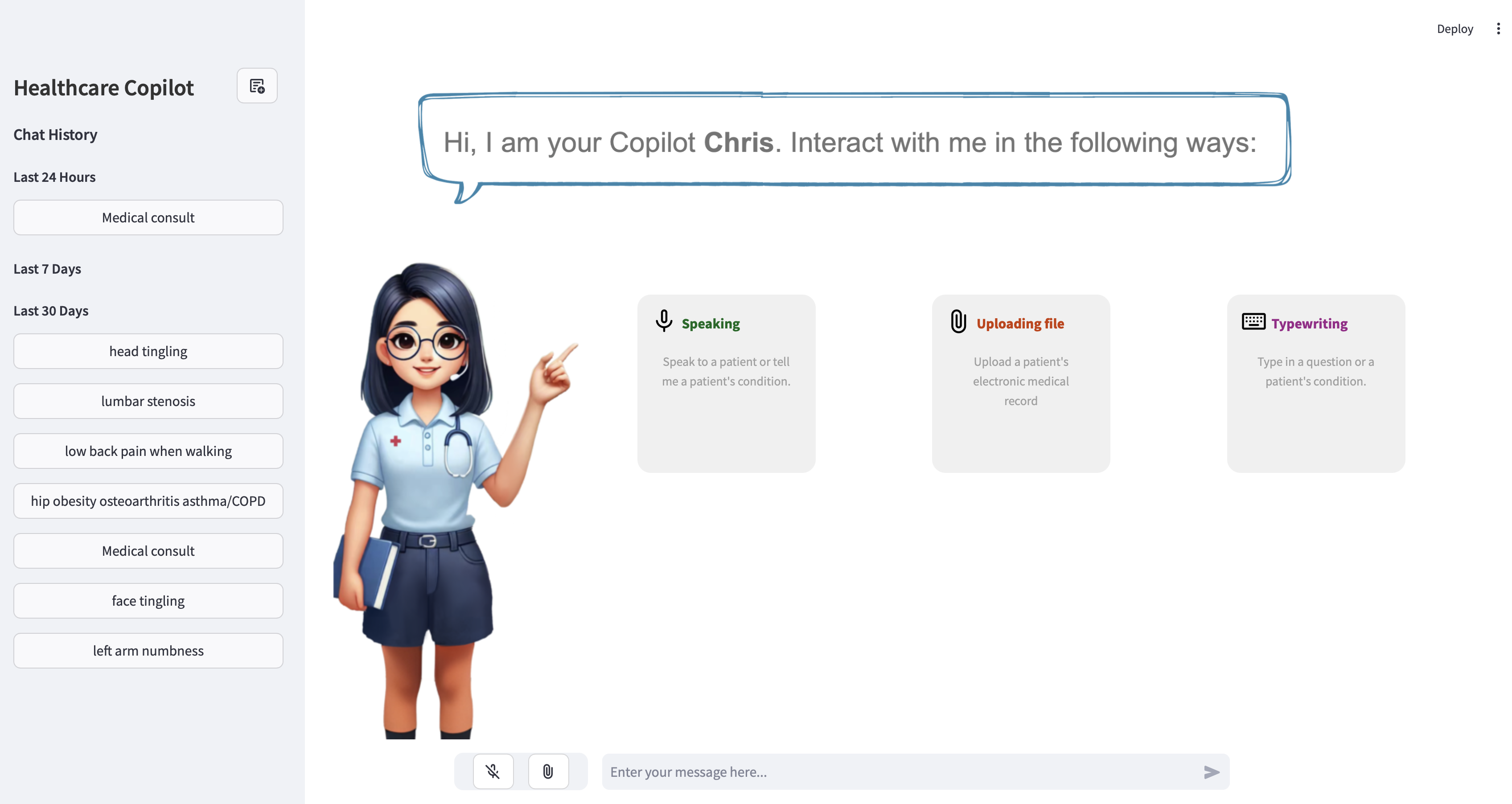} 
    \caption{User Interface of MedRAG} 
    \label{fig:UI} 
\end{figure}

We provide a user-friendly interface built with Streamlit and CSS, designed to facilitate interaction with MedRAG’s diagnostic module. As shown in Figure~\ref{fig:UI}, the left panel displays the chat history, allowing quick access to past consultations. The main panel presents three input modes: Speaking, Uploading Files, and Typewriting. The bottom section provides a text input field for direct user queries.

\subsection{Case Study}

\vspace{-5pt}
\begin{table}[htb]
    \centering
    \small
    \renewcommand{\arraystretch}{1} 
    \begin{tabular}{p{2.0cm} p{5.8cm}}
        \toprule
        \textbf{System} & \textbf{Diagnostic Suggestion} \\  
        \midrule
        \textbf{Query} & Provide diagnosis suggestions for the following patient: Age: 47.  
        Functional status: Difficulty walking [...]
        Description: Pain from right lower back radiates down to buttock and right posterior lower limb. \\
        \midrule
        \textbf{Llama3.1-8b} & Lumbar Radiculopathy, Sciatica, [...]. \\
        \textbf{Mixtral-8x7b} & It is possible that the patient is experiencing pain due to sciatica. \\
        \textbf{Qwen2.5-72b} & Potential Diagnoses: Sciatica [...];  
        Lumbar Herniated Disc [...]; Spinal Stenosis: [...]. \\
        \midrule
        \textbf{MedRAG (Ours)} &  Lumbar canal stenosis.
        
        You can further ask:  Is the pain worse when standing or walking down hill? \\
        \bottomrule
    \end{tabular}
    \vspace{-3pt}
    \caption{Comparison of Diagnostic Suggestions Across Systems}
    \label{tab:diagnosis_comparison}
\end{table}
\vspace{-8pt}

In Table~\ref{tab:diagnosis_comparison}, we compare MedRAG with other LLMs including Llama3.1-8b, Mixtral-8x7b and Qwen2.5-72b, which often provide incorrect or ambiguous diagnoses, such as sciatica or radiculopathy, and struggle to distinguish similar conditions. In contrast, MedRAG accurately identifies lumbar canal stenosis and proactively generates follow-up questions to help doctors further refine the diagnosis.

Furthermore, a detailed demonstration scenario of a medical consultation with the corresponding suggested follow-up question, along with an end-to-end evaluation including voice modality, are provided in Appendix A-B\footnotemark and Table~\ref{mod}.
\footnotetext{\tiny\textit{https://github.com/SNOWTEAM2023/MedRAG/tree/main/appendix}}

\vspace{-6pt}
\vspace{-6pt}

\vspace{-6pt}
\begin{table}[h]
\centering
\small
\begin{tabular}{lcccc}
\toprule
\textbf{Backbone LLM} & \textbf{Modal} & \textbf{L1} & \textbf{L2}& \textbf{L3} \\
\midrule
GPT-4o   & text   & 91.87 & 81.78 & 73.23 \\
GPT-4o   & voice  & 88.23 & 78.43 & 70.58 \\
GPT-3.5-turbo   & text   & 70.56 & 68.68 & 50.57 \\
GPT-3.5-turbo   & voice  & 64.70 & 60.78 & 45.09 \\
\bottomrule
\end{tabular}
\vspace{-3pt}
\caption{Evaluation of Different Modal on CPDD}
\label{mod}
\end{table}
\vspace{-8pt}

\subsection{Doctor Evaluation}

To complement quantitative benchmarks with clinical insights, we incorporated a human evaluation involving four experienced doctors. These experts' feedback provides essential perspective on how MedRAG is perceived in clinical contexts, particularly in terms of trust and usability.

For the evaluation, doctors assessed three representative test cases with responses from both MedRAG and GPT-4o, focusing on functional design, user interface, EHR analysis, and medical consultation analysis. Our evaluation uses five Human Factors criteria (e.g., Clinical Relevance and Trust) widely used to assess AI-assisted systems~\cite{choudhury2023investigating,choudhury2022factors,zhao2021brain}. The details of the criteria definitions and specific questions are provided in Appendix C~\footnotemark[\value{footnote}].





\vspace{-7pt}
\begin{figure}[h] 
    \centering
    \includegraphics[width=0.8\linewidth]{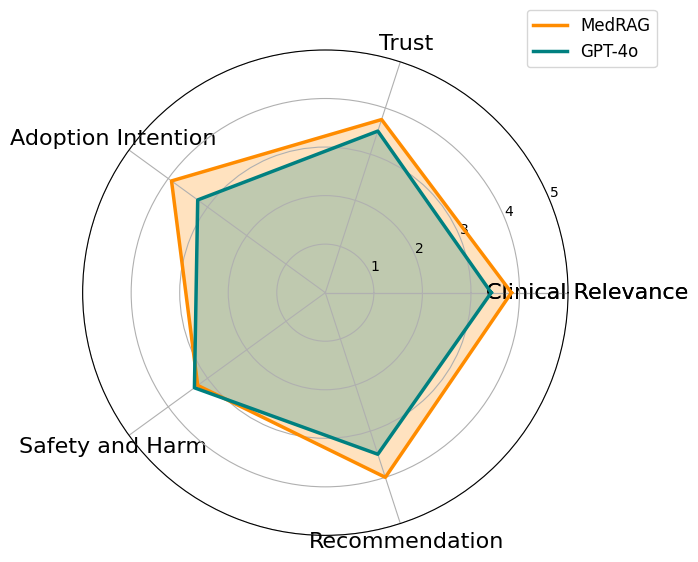} 
    \caption{Result of Doctor Evaluation} 
    \label{fig:evaluation_radar} 
\end{figure}
\vspace{-7pt}

The comparative results are presented in Figure~\ref{fig:evaluation_radar}. The results demonstrate that MedRAG outperforms GPT-4o across all criteria, with particularly outstanding performance in Adoption Intention. Some doctors emphasized that, since evidence-based practice is fundamental to medicine~\cite{prasad2014fundamentals}, MedRAG stood out for its strong emphasis on evidence-based reasoning.

\section{Conclusion}
MedRAG is a smart multimodal healthcare copilot with powerful LLM reasoning, integrating multimodal inputs and KG-elicited reasoning to enhance diagnostic accuracy and decision support. The results of case studies and doctor evaluation have consistently demonstrated the effectiveness and reliability of MedRAG in medical decision-making contexts.

\newpage
\section*{Acknowledgments}
    This research is supported by the Joint NTU-UBC Research Centre of Excellence in Active Living for the Elderly (LILY) and the College of Computing and Data Science (CCDS) at NTU Singapore. It is also partially supported by the Singapore Ministry of Education Academic Research Fund Tier 1 (Grant No. 2017-T1-001-270). This research is also supported, in part, by the National Research Foundation, Prime Minister’s Office, Singapore under its NRF Investigatorship Programme (NRFI Award No. NRF-NRFI05-2019-0002). Any opinions, findings, conclusions, or recommendations expressed in this material are those of the authors and do not reflect the views of National Research Foundation, Singapore. This research is supported, in part, by the Singapore Ministry of Health under its National Innovation Challenge on Active and Confident Ageing (NIC Project No. MOH/NIC/HAIG03/2017). This research is supported, in part, by the RIE2025 Industry Alignment Fund – Industry Collaboration Projects (IAF-ICP) (Award I2301E0026), administered by A*STAR, as well as partially supported by Alibaba Group and NTU Singapore through Alibaba-NTU Global e-Sustainability CorpLab (ANGEL). This work is partially supported by the Wallenberg Al, Autonomous Systems and Software Program (WASP) funded by the Knut and Alice Wallenberg Foundation.

\bibliographystyle{named}
\bibliography{ijcai25}

\end{document}